\documentclass{article}
\pdfoutput=1

\PassOptionsToPackage{numbers, compress}{natbib}

\usepackage[preprint]{neurips_2022}
\PassOptionsToPackage{numbers, compress}{natbib}
\usepackage{microtype}
\usepackage{graphicx}
\usepackage{wrapfig}
\usepackage{subcaption}
\usepackage{float}
\usepackage{amsmath}
\usepackage{cancel}
\usepackage{tikz}
\usetikzlibrary{bayesnet}
\usetikzlibrary{matrix}
\usepackage{booktabs} 

\usepackage{hyperref}

\title{Improving VAE-based Representation Learning}
\author{%
   Mingtian Zhang\textsuperscript{1} \And Tim Z. Xiao\textsuperscript{2} \And Brooks Paige\textsuperscript{1}\And David Barber\textsuperscript{1}
   \Aff
  \textsuperscript{1}Centre for Artificial Intelligence, University College London\\
    \textsuperscript{2}University of Tübingen \& IMPRS-IS\\
  \textsuperscript{1}\texttt{\{m.zhang,b.paige,d.barber\}@cs.ucl.ac.uk} \\
  \textsuperscript{2}\texttt{zhenzhong.xiao@uni-tuebingen.de} \\
}

\begin{document}
\maketitle
\begin{abstract}

Latent variable models like the  Variational Auto-Encoder (VAE) are commonly used to learn representations of images. 
However, for downstream tasks like semantic classification, the representations learned by VAE are less competitive than other non-latent variable models. 
This has led to some speculations that latent variable models may be fundamentally unsuitable for representation learning.
In this work, we study what properties are required for good representations and how different VAE structure choices could affect the learned properties. 
We show that by using a decoder that prefers to learn local features, the remaining global features can be well captured by the latent, which significantly improves performance of a downstream classification task. 
We further apply the proposed model to semi-supervised learning tasks and demonstrate improvements in data efficiency.

\end{abstract}


\section{Introduction}
Finding good representations is a crucial but challenging step in many machine workflows  \citep{bengio2013representation}. 
Many methods have been proposed to learn better representations for different applications.
In natural language processing,  unsupervised pre-training on language modeling \citep{brown2020language, radford2018improving, devlin2018bert} has shown promising improvement for many downstream tasks such as machine translation \citep{zhu2020incorporating}, sequence labeling \citep{shah2021locally}, and question answering \citep{zhu2021retrieving}. 
Similarly, in computer vision, self-supervised techniques has been used for creating various state-of-the-art visual representations to improve image classifications \citep{chen2020simple,he2020momentum, caron2019unsupervised, oord2018representation}, as well as object detection \citep{xie2021detco} and instance segmentation \citep{desai2021virtex}. 



From a modeling perspective, a natural model family for learning representations is the latent variable model. For example,  Variational Auto-Encoder (VAE)~\cite{kingma2013auto,rezende2014stochastic} is a popular latent variable model parameterized by non-linear neural networks. Despite its big success in applications like image generation \cite{razavi2019generating,vahdat2020nvae} and lossless compression \cite{townsend2019practical,townsend2019hilloc}, VAE is less competitive in the representation learning tasks comparing to other non-latent variable models~\cite{hjelm2018learning,oord2018representation}. This has led to some speculations that latent variable models may be fundamentally unsuitable for representation learning. Therefore, we are interested in studying  the influencing factors of the VAE-based representation learning towards improving the downstream task performance. Our results show that by incorporating the right inductive bias in the model structures, the representations that learned by VAEs can achieve better down-stream task performance comparing to other popular representation learning schemes. 


\section{Representation Learning with Variational Auto-Encoders}
Given a dataset $\mathcal{X}=\{x^1,\ldots,x^N\}$ sampled identically and independently (i.i.d.) from an underlying data distribution $p_d(x)$, we want to learn a latent variable model $p_\theta(x)=\int p_\theta(x|z)p(z)dz$ to approximate $p_d(x)$.
The parameter $\theta$ is usually trained by maximizing the likelihood  $\frac{1}{N}\sum_{n=1}^N \log p_\theta(x^n)$.
When $\theta$ is parameterized by a neural network, the evaluations of the log likelihood $\log p_\theta(x)$ is usually intractable. Instead, the evidence lower bound (ELBO) can be used to train the model
\begin{align}
    \log p_\theta(x)&\geq \langle \log p_\theta(x|z)\rangle_{q_{\phi(z|x)}}-\mathrm{KL}(q_\phi(z|x)||p(z))\equiv \mathrm{ELBO}(x,\theta,\phi),
\end{align}
where we use $\langle\cdot\rangle$ to denote integration, i.e.\ $\langle f(x)\rangle_{p(x)}=\int f(x)p(x)dx$. This model is referred to as the Variational Auto-Encoder (VAE) \cite{kingma2013auto,rezende2014stochastic}, where the  amortized posterior or ``encoder'' $q_\phi(z|x)$ is introduced to approximate the true posterior $p_\theta(z|x)\propto p_\theta(x|z)p(z)$. Therefore, the learned $q_\phi(z|x)$ can be used to generate the representation. For of a given data $x'$, common ways of obtaining a representation includes sampling from the amortized posterior $z'\sim q_\phi(z|x')$, finding the  most likely representation $z'=\arg\max q_\phi(z|x')$ \cite{bengio2013representation} or using an embedding of the distribution $q(z|x')$ as the representation~\cite{sriperumbudur2010hilbert,gretton2012kernel}. In Section \ref{sec:different:representations}, we discuss the properties of different representation types and empirically study the practical affects to the down-stream tasks.

\section{What Makes a Good Representation?}
For any representations extracted by a function that loses information, e.g. a non-invertible encoder maps from a high-dimensional data space to a low-dimensional representation space, a downstream task can always be designed to be based on the lost information and can then have arbitrary bad performance. Therefore, the concept of ``universal'' representation learning is ill-defined. In this work, we are interested in the down-stream classification task, which is one of the most popular use cases of representation learning~\cite{bengio2013representation,van2018representation}. We then discuss the desired properties of the representations for the focused task and the corresponding evaluation metrics to verify these properties.






A valid representation should contain sufficient information for the downstream classification labels. 
However, the \emph{sufficiency} property alone is not enough to guarantee a good representation.
For example, the original data $x$ itself or any invertible transformations of $x$  will have sufficient information, but they  also contain other redundant information that is irrelevant to the downstream labels. Therefore, another natural requirement is that, while preserving sufficient information about the labels, the representations should contain \emph{minimal} information about the data \cite{dubois2020learning}.
The concept of combining \emph{sufficiency} and \emph{minimality} is widely used in different areas, e.g. minimal sufficient statistics \cite{wasserman2004all} or rate-distortion theory \cite{cover1999elements}. It is also the fundamental intuition behind  \emph{information bottleneck principle} \cite{tishby2000information,shamir2010learning,tishby2015deep,dubois2020learning}.  We then introduce several metrics that can reflect these two properties in the VAE-based representation learning scenario.

\textbf{Nonlinear Probe} The \emph{sufficiency} property can be evaluated by fitting a nonlinear neural network classifier to the representations since it is able to extract any information of the input data if we assume a flexible network parameterization. This evaluation method is referred to as  \emph{nonlinear probe}.

\textbf{Mutual Information} The \emph{minimality} can be measured by the mutual information between the  data $x$ and its representation $z$, which is formally defined as
\begin{align}
    \mathrm{I}(X_d,Z)\equiv\left\langle\log\frac{p_d(x)p_\theta(z|x)}{p_d(x)p(z)}\right\rangle_{p_d(x)p_\theta(z|x)},
\end{align}
where $X_d$ is the data random variable and the true posterior $p_\theta(z|x)$ can be approximated using $q_\phi(z|x)\approx p_\theta(z|x)$, thus the mutual information can be approximated as
\begin{align}
    \mathrm{I}(X_d,Z)&\approx \left\langle\log\frac{p_d(x)q_\phi(z|x)}{p_d(x)p(z)}\right\rangle_{p_d(x)q_\phi(z|x)}= \Big\langle\mathrm{KL}(q_\phi(z|x)||p(z))\Big\rangle_{p_d(x)},\label{eq:mi}
\end{align}
where we can use the test data to conduct a Monte-Carlo approximation of the integral $\langle\cdot\rangle_{p_d(x)}$.

\textbf{Intrinsic Dimension} An alternative perspective of \emph{minimality} is that the representations should lie on a low dimensional manifold. However, since the ambient dimension of the representations is pre-fixed before training, the \emph{minimality} can be reflected by the \emph{intrinsic dimension} of the representations. In this paper, we estimate the intrinsic dimension by applying a PCA\footnote{We assume the representations lie on a linear subspace. There are other nonlinear intrinsic dimension estimation is available, e.g. \cite{zhang2021flow}, we leave that to future exploration.} on the representations and using the number of the non-zero eigenvalues as the intrinsic dimension.

\textbf{Linear Probe} 
    Another common evaluation method is using a linear classifier (e.g. a linear SVM) for classification, which is also called \emph{linear probe}~\cite{alain2016understanding, hjelm2018learning,bengio2013representation,ma2020decoupling,oord2018representation}. Intuitively, a representation that doesn't contain too much information about the data will have bad performance with linear probe, which indicates the linear probe can reflect the \emph{sufficiency} to some extent.
Recent works \cite{gorban2018correction,gorban2020high} also shows the following  connection: the linear separability increases when the intrinsic dimension of the input features decreases, which suggests the linear probe is affected by the intrinsic dimension of the representations. Additionally, \cite{tishby2015deep}  shows that the linear separability has a close relationship to the information bottleneck curve, which indicates that the linear probe can reflect both sufficiency and minimality, while the nonlinear probe can only reflects sufficiency. 

In the following sections, we consider the image classification task and discuss how different VAE structures can affect the sufficiency and minimality of the learned representations.



\section{Leanring Representations for Image Classification}
\begin{wrapfigure}{r}{0.4\textwidth}
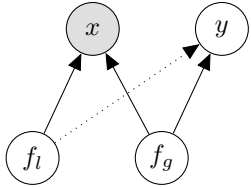

\vspace{-0.3cm}
\centering
\tikz{
         \node[obs] (x1) {$x$};
          \node[latent, below=of x1,, xshift=-0.8cm] (z1) {$f_l$};
          \node[latent, right=of z1] (z2) {$f_g$};
           \node[latent, right=of x1] (z3) {$y$};
         \edge {z1} {x1};
          \edge {z2} {x1};
          \edge {z2} {z3};
    \draw [ dotted , ->] (z1) -- (z3);
    }
\caption{Graphical model  of the classification assumption, where $f_g,f_l$ denote global and local feature respectively and $y$ denotes the class label. The dashed line indicates the weak dependency.\label{fig:data:assumption}}
\vspace{-0.1cm}
\end{wrapfigure}

The features of  an image can be generally divided into low-level  and high-level categories \cite{szeliski2010computer}.  Low-level features usually contain  properties like color, texture or edges and corners \cite{marr1982vision}. These features can be extracted from a local patch of images, which are also referred to as the local features \cite{lowe2004distinctive,bay2006surf,yi2016lift}. High-level (or global) features are composed by local features and contains the semantic information~\cite{jegou2011aggregating,hjelm2018learning,ma2020decoupling}. We are interested in the semantic-level image classification, where  the class of a image is strongly depends on its global features and weakly depends on its local features, see Figure \ref{fig:data:assumption} for an illustration. This assumption has also been implicitly used in many representation learning works \cite{chen2020simple,he2020momentum,noroozi2016unsupervised}. However, this assumption may not hold for the datasets where labels depend on the local features. For example, in  medical radiology, 
the clinically useful information is usually contained in highly
localized regions~\cite{shyu1998local}, which is out of the scope of this work.


We are then ready to study how the representations learned by different VAE structures will affect the down-stream  semantic image classification.
We notice that the representations of the VAE depends on the amortized posterior (encoder) $q_\phi(z|x)$, whose optima is uniquely determined by the decoder $p_\theta(x|z)$ and the prior $p(z)$: $q_{\phi^*}(z|x)\propto p_\theta(x|z)p(z)$\footnote{Although in practice, there is usually a gap between $q_{\phi^*}(z|x)$ and $p_\theta(x|z)$, see \cite{zhanggeneralization,cremer2018inference,shu2018amortized} for a detailed introduction and the corresponding improving methods.}.  Therefore, the choice of the decoder structure plays a key role in learning the representations.
Classic VAE assumes a conditional independent decoder, whereas other decoder variants, e.g. VAE with an autoregressive decoder~\cite{gulrajani2016pixelvae}, has also been proposed to
improve the image generalizations. However, the effects of using such a decoder to the representation learning remain under-explored, we thus give a detailed discussion below.

\subsection{VAE with a Conditional Independent Decoder}
\begin{wrapfigure}{r}{0.5\textwidth}
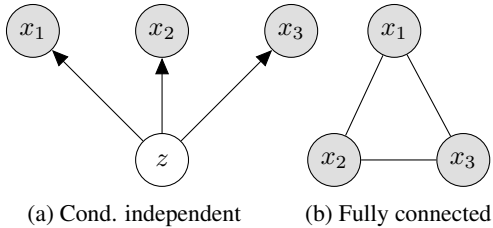

\vspace{-0.3cm}
\centering
\begin{minipage}[c]{.49\linewidth}
\centering
\tikz{
         \node[obs] (x1) {$x_1$};
          \node[obs, right=of x1] (x2) {$x_2$};
          \node[obs, right=of x2] (x3) {$x_3$};
           \node[latent, below=of x2] (z) {$z$};
         \edge {z} {x1};
          \edge {z} {x2};
          \edge {z} {x3};
}
\subcaption{Cond. independent}
\end{minipage}
  \begin{minipage}[c]{.49\linewidth}
  \centering
\tikz{
 \node[obs] (x1) {$x_1$};
 \node[obs, below=of x1, xshift=-0.8cm] (x2) {$x_2$};
  \node[obs, right=of x2] (x3) {$x_3$};
 \edge[-] {x1} {x2};
 \edge[-] {x2} {x3};
 \edge[-] {x3} {x1};
}
\subcaption{Fully connected\label{fig:connected}}
\end{minipage}
\captionof{figure}{ Figure a shows the pixels are conditionally independent given the latent $z$. Figure b shows the pixel random variables are fully connected when integrating over $z$. \label{fig:cond:ind}}
\vspace{-0.1cm}
\end{wrapfigure}
For image modelling with a classic VAE, the decoder usually has a conditional independent structure~\cite{kingma2013auto}.
Specifically, for an image with dimension $I\times J$, the latent variable model is $p_\theta(x)=\int \prod_{ij} p_\theta(x_{ij}|z)p(z)dz$,
where each $x_{ij}$ is conditionally independent given the latent $z$. 
After integrating out the latent variable $z$, all $x_{ij}$ become fully connected, so all the correlations between pixels are modeled through the latent variable $z$, see Figure \ref{fig:connected}  for a graphical model illustration.
For a simple data distribution (e.g. MNIST) that can be well-approximated by a VAE with the conditional independent decoder,  the latent representation will contain all the correlation features between pixels within the images, which includes both local features and global features. In this case, the representations will satisfy the \emph{sufficiency} requirement. 

However, for a complex distribution like CIFAR10, the conditional independent VAE with a small latent size (e.g. 64, which is commonly used in the literature \cite{ma2020decoupling,hjelm2018learning}) is  insufficient to learn a good approximation of the data distribution.  In this case, the latent is no longer  able to capture all the correlations between pixels and either local or global features can be lost during training,  which degrades the \emph{sufficiency} property of the representations.
 Since the global features dominate performance of the downstream classification task by the assumption in Figure~\ref{fig:data:assumption},
 one solution  is to use a decoder that is capable of learning local features and leaving the remaining global features to be captured by the latent.  We then discuss how an autoregressive  decoder can help achieve this goal.

\subsection{VAE with an Autoregressive Decoder}
 In nature image, two nearby pixels are usually very similar and have stronger correlations than  pixels that are far away from each other, so the likelihood of image models is usually dominated by the local features \cite{schirrmeister2020understanding,zhang2021out}.
 However, the conditional independence VAE is blind to this fact and solely relying  on the latent $z$ to capture all kinds of correlations, which  leads to a  low test likelihood on images. By contrast, autoregressive models like PixelCNN \citep{gulrajani2016pixelvae,salimans2017pixelcnn++}
 can naturally capture this inductive bias
 \begin{align}
     p_\theta(x)=\prod_{ij} p_\theta(x_{ij}|x_{[1:i-1,1:J]},x_{[i,1:j-1]}),
 \end{align}
 where we denote $[x_{[1:i-1,1:J]},x_{[i,1:j-1]}]\equiv x^{past}_{ij}$ and $p_\theta(x_{11}|x_{11}^{past})=p_\theta(x_{11})$. The  PixelCNN can be implemented by stacking several masked convolution layers
\citep{gulrajani2016pixelvae} with kernel size $k\times k$ (where $k>1$). Therefore, when the depth of the layers increases, the dependency horizon also scales up towards a fully autoregressive model. Comparing to the conditional independent decoder, it is more easy for the autoregressive to capture the local dependency, which results in higher likelihoods.

However, the autoregressive models doesn't allow a low-dimensional  representation of the data. A natural idea is to  use an autoregressive decoder \citep{gulrajani2016pixelvae} in the latent variable model
\begin{align}
    p_\theta(x)&=\int p(z)\prod_{ij} p_\theta(x_{ij}|x^{past}_{ij},z)dz.\label{eq:pixelvae}
\end{align}
Similar to VAE training, a lower bound of the log likelihood can be constructed for training the model
 \begin{align}
\log p_\theta(x)&\geq \big\langle\sum_{ij}\log p_\theta(x_{ij}|x_{ij}^{past},z)\big\rangle_{q_\phi(z|x)} -\mathrm{KL}(q_\phi(z|x)||p(z)).
\end{align}
We refer to this model as the \emph{Full PixelVAE (FPVAE)}.
 In principle, a flexible autoregressive decoder can have the ability to capture all the information within the images which includes both local and global features. When this happens, the decoder does not depend on the latent anymore and the first term in the training objective collapses to
  \begin{align}
&\big\langle\sum_{ij}\log p_\theta(x_{ij}|x_{ij}^{past},\cancel{z})\big\rangle_{q_\phi(z|x)} \longrightarrow \sum_{ij}\log p_\theta(x_{ij}|x_{ij}^{past}).\label{eq:fpvae:elbo}
\end{align}
The remaining KL term in Equation \ref{eq:fpvae:elbo} will drive $q_\phi(z|x)$ to be close to $p(z)$, makes the learned representations uninformative. This phenomenon is  also referred to as latent collapse~\cite{he2019lagging,lucas2019understanding}. However,  for a PixelCNN decoder with no BatchNorm~\cite{ioffe2015batch}, the latent collapse phenomenon doesn't happen during training \cite{gulrajani2016pixelvae}, since it's more difficult for the PixelCNN to capture long-term dependency comparing to a latent variable model. In this case, the decoder prefers to learn local features, leaving the global features to be captured by the latent. However, the decomposition of the local and global features learned by two parts of the model is not transparent in the FPVAE. To further investigate the information separation procedure, we propose to use a local autoregressive model as the decoder, which allows us to explicitly control the scale of the local dependency and therefore limit the information learned by the decoder, see the following section for an introduction.


\subsection{VAE with a Local Autoregressive Decoder}
\begin{wrapfigure}{r}{0.3\textwidth}
    \centering
    \includegraphics[width=0.6\linewidth]{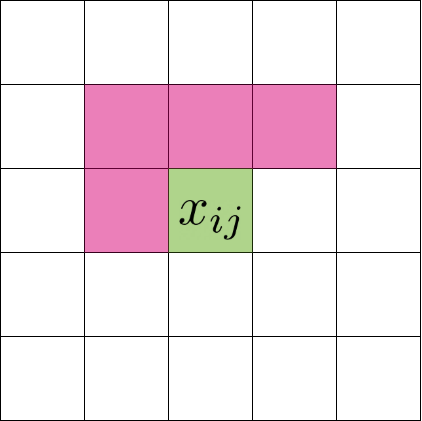}
    \captionof{figure}{Local dependency  horizon $h=1$, pixel $x_{ij}$ only depends on the pink pixels within a local region.}
    \label{fig:local:dependency}
    \vspace{-0.7cm}
\end{wrapfigure}

We propose to use the local autoregressive model \citep{zhang2021out,zhang2022parallel} as the decoder, the model can be written as
 \begin{align}
 \resizebox{0.4\hsize}{!}{$
  p_\theta(x)=\int p(z)  \prod_{ij}p_\theta(x_{ij}|x^{local}_{ij},z)dz,$}
  \end{align}
  where  $x_{ij}^{local}=x_{[i-h:i-1,j-h:j+h]},x_{[i,j-h:j-1]}$ and $h$ denotes the dependency horizon of $x_{ij}$. Figure \ref{fig:local:dependency} illustrates the dependency structure with $h=1$. In practice, we pad the images using 0 with width $h$ to prevent cases like $i<h$ or $j<h$. Paper \cite{zhang2021out} proposes to implement local PixelCNN with dependency horizon $h$ by letting  the first masked convolution layer
 of the PixelCNN has kernel size $k\times k$, 
where $k=2h+1$, and  other subsequent  layers has $1\times 1$ kernels. 
Alternatively, one can stack $h$ masked convolution kernels with size $3\times 3$, followed by $1\times 1$ convolution layers, which gives same dependency horizon $h$ and is more flexible. We refer to the VAE with a local PixelCNN decoder as the \emph{Local PixelVAE} (LPVAE).

By varying the dependency horizon length, we can control the decoder's ability of learning the local features, thereby controlling the amount of global information that is remained to be captured by the latent representations. We are now ready to use the LPVAE model family and the proposed metrics to study how local and global features  affect the properties of the learned representations in practice.

\section{Empirical Studies of VAE-based Representation Learning}
We empirically study the factors that affect the representations that learned by VAE\footnote{The code of the experiments can be found in the following link: \url{https://github.com/zmtomorrow/ImprovingVAERepresentationLearning}, all the results are conducted on a NVIDIA Tesla V100 GPU.}. In Section 5.1, we compare three different types of representation that can be obtained from the encoder. In Section 5.2 and 5.3, we study  how different decoder structures affect the properties of the learned representations. Especially, we focus on  two scenarios where the training data  can (MNIST~\cite{lecun1998mnist}) and cannot (CIFAR10~\cite{krizhevsky2009learning}) be well approximated by a conditional independent VAE.
In all experiments, we use a linear SVM as the linear probe and the nonlinear probe is a two-layer linear net with hidden size 200 and ReLU activation, the BatchNorm and dropout (with rate 0.1)  are also used in the network. The linear and nonlinear probe methods are the same as that used in \cite{ma2020decoupling,hjelm2018learning}. 

\subsection{Types of the VAE Representation\label{sec:different:representations}}
As discussed in Section 2, there are  three types of representations can be obtained for a given data $x'$:
\begin{table*}
\begin{center}
     \begin{tabular}{cccccc}\hline
       Model & VAE ($h=0$) & LPVAE ($h=1$) & LPVAE ($h=2$) & FPVAE \\ \hline
         
        Conditional Entropy & -3.9 & 25.4 & 32.8 & 34.6\\\hline
        Sample (k=1) & 97.3$\pm$0.2  & 94.5$\pm$ 0.2 & 90.5$\pm$0.2 & 89.3$\pm$0.3 \\
        Sample (k=100) & 97.8$\pm$0.1  & 97.4$\pm$ 0.0 & 96.0$\pm$0.0 & 95.6$\pm$0.1\\
        Max. a posteriori & 97.9$\pm$0.1  & 97.8$\pm$0.1 & 96.9$\pm$0.1 &96.8$\pm$0.1\\
        Dist. embedding & 98.0$\pm$0.2 & 97.9$\pm$0.1 & 97.1$\pm$0.1 & 97.1$\pm$0.1\\

         \hline
\end{tabular}

\end{center}
\caption{Comparisons between different representation methods on MNIST classification task. The reported four VAE models share the same encoder structure but have $q_\phi(z|x)$ with different conditional entropy. The model specifications can be found in Section 5. \label{tab:det:vs:sto}}
\end{table*}

1. \textbf{Posterior sampling} Representations of $x'$ can be the samples $z'\sim q_\phi(z|x')$. For the downstream classification task, the predictive distribution $p(y|x)=\int p(y|z)q_\phi(z|x)dz$ is approximated by  Monte-Carlo: $p(y|x)\approx \frac{1}{K}\sum_{k=1}^K p(y|z_k)$, where $z_k\sim q_\phi(z|x)$. For a distribution $q_\phi(z|x)$ with large entropy $\mathrm{H}(q_\phi(z|x))$, a large number of samples needs to be used to obtain a good approximation.

2. \textbf{Maximum a posteriori} The MAP estimation  $z'=\arg\max_{z} q_\phi(z|x')$ is commonly used as the representation~\cite{bengio2013representation}. For a Gaussian amortized posterior  $\mathcal{N}(\mu_\phi(x'),\sigma^2_\phi(x'))$, the representation will be  $z'=\mu_{\phi}(x')$. This scheme is computationally efficient since it doesn't need Monte-Carlo integration and the dimensional of the representation is equal to the latent dimension $\mathrm{Dim}(Z)$.
    
3. \textbf{Distribution embedding} We can also use a deterministic vector to represent a distribution~\cite{sriperumbudur2010hilbert,gretton2012kernel}. For a Gaussian $q_\phi(z|x')$, a simple embedding is a concatenation of its mean and standard deviation (std), which creates an one-to-one mapping between the posterior distribution and a vector. This representation requires a larger vector dimension ($2\times \mathrm{Dim}(Z)$) comparing to the MAP representation. 

We  compare the three types of representations with four VAEs: VAE with a conditional independent decoder (which is a special case of the LPVAE with $h=0$); LPVAE with $h=1$ and $h=2$ and a FPVAE. All models have  $\mathrm{Dim}(Z)=32$ and share the same encoder structure: 3 linear layers with 500 hidden units with BatchNorm and ReLU activation. Each decoder  contains 3 linear layers with output channel 32 and a (local) PixelCNN module with 2 masked CNN layers followed by 5 Residual blocks \cite{he2016deep}. For the local PixelCNN, the first CNN layer has kernel size is $k\times k$ where $k=2h+1$ and other subsequent  kernels have sizes $1\times 1$. For the full PixelCNN, the subsequent kernels have size $3\times3$ to increase dependency horizon when stacking multiple masked CNN layers. All models are trained for 100 epochs with batch size 100 and  $\text{lr}=3\times 10^{-4}$ using Adam~\cite{kingma2014adam} optimizer. 

 We train the models on the grayscale MNIST and obtain four encoders with different conditional entropy (defined as $\int p_d(x)\mathrm{H}(q_\phi(z|x))dx$). 
We then fit a 2 layer neural network on the representations to learn a classifier $p(y|z)$ for each VAE.  Table \ref{tab:det:vs:sto} shows the test classification accuracy for three kinds of representation.  For posterior sample representation, we show the results with sample number $k=1$ and $k=100$. We can find  when the conditional entropy of the $q_\phi(z|x)$ becomes larger, the Monte Carlo approximation with 1 sample will become worse comparing to the one using 100 samples. The distribution embedding representation achieves the best performance among the three methods but requires the dimension of the representation to be 64. On the other hand, the MAP representation is slightly worse than the distribution embedding method but better than sampling methods and only requires representation dimension 32. Nevertheless, since the difference between different methods are marginal ($\leq 1.5\%$)
, we will focus on the MAP representation in this paper, which has the best computational efficiency and accuracy trade-off in this demonstration example.
This representation is also the most commonly used scheme in the literature \cite{bengio2013representation,ma2020decoupling}.

\subsection{Representation Learning on MNIST} \label{mnist_exp}
We then study how different VAE structures will affect the learned  representations using four different VAEs (LPVAE with $h=\{0,1,2\}$ and a FPVAE)  that discribed in the previous section.
Table \ref{tab:mnist} shows the test BPD\footnote{Bits-per-dimension (BPD) represents the negative $\log_2$ likelihood normalized by the data dimension. Lower BPD indicates higher likelihood.} of the models. We can find the conditional independent VAE well approximates the MNIST and achieves a decent BPD (1.24). When the decoder's dependency horizon is increased, the test BPD also goes down, which suggests that autoregressive model can better capture the local features in the images since the likelihood is dominated by the local features \cite{schirrmeister2020understanding,zhang2021out}.

\begin{table}[t]
\caption{Representation learning on MNIST, both probe results are calculated over 3 random seeds.\label{tab:mnist}}
\centering
 \begin{tabular}{cccccc}\hline
      Model & VAE ($h=0$) & LPVAE ($h=1$) & LPVAE ($h=2$) & FPVAE \\
        BPD & 1.24 &  1.01 & 1.02 & \textbf{0.98}\\ \hline 
        Mutual Info & 49.8 & 20.1 & 12.3 &10.7\\
        Intrinsic Dim & 25 & 15 & 9 & 9\\ \hline
        Nonlinear Probe& \textbf{97.9}$\pm$0.1  & 97.8$\pm$0.1 & 96.9$\pm$0.1 &96.8$\pm$0.1\\
        Linear Probe & 90.9$\pm$0.1  & 94.4$\pm$0.1 & \textbf{95.2}$\pm$0.1 & \textbf{95.2}$\pm$0.1 \\ \hline
\end{tabular}
\end{table}
\begin{figure}[t]
\vspace{-0.2cm}
\centering
    \begin{subfigure}[t]{0.23\linewidth}
    \centering
        \includegraphics[width=\textwidth]{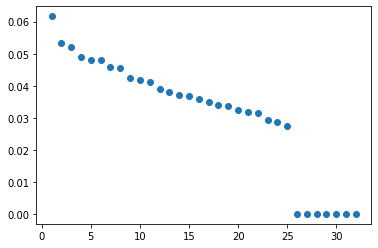}
        \caption*{{\footnotesize	 VAE ($h=0$)}}
    \end{subfigure}
    \begin{subfigure}[t]{0.23\linewidth}
    \centering
        \includegraphics[width=\textwidth]{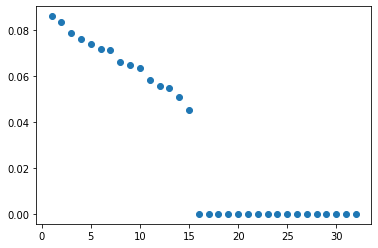}
        \caption*{{\footnotesize	 LPVAE ($h=1$)}}
    \end{subfigure}
    \begin{subfigure}[t]{0.23\linewidth}
    \centering
        \includegraphics[width=\textwidth]{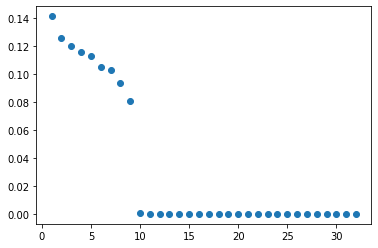}
        \caption*{{\footnotesize	 LPVAE ($h=2$)}}
    \end{subfigure}
    \begin{subfigure}[t]{0.23\linewidth}
    \centering
        \includegraphics[width=\textwidth]{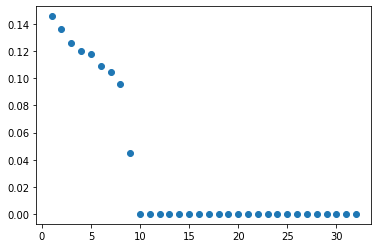}
        \caption*{{\footnotesize	 FPVAE}}
    \end{subfigure}
    \caption{Eigenvalues (sorted from high to low) of the representations learned by four VAE variants. The numbers of  non-zero eigenvalues indicate the intrinsic dimensions of the representations. \label{fig:mnist:pca}}
\end{figure}

As we discussed in Section 4.1, when a conditional independent VAE that well-approximates the data distribution, the learned representations will contain both local and global features.
Table \ref{tab:mnist} shows the representations learned by the conditional independent VAE achieves highest nonlinear probe accuracy comparing to other VAE variants, which validates the relationship between sufficiency and non-linear probe.
When we increase the dependency horizon  from 0 to 2, the autoregressive decoder has more flexibility to capture local features and the remaining information learned by the latent decreases, which is revealed by the decreasing mutual information and intrinsic dimensions, see Figure \ref{fig:mnist:pca}.  We find that losing local information in the latent only results in a marginal decrease of the nonlinear probe, which is consistent with our assumption  (Figure \ref{fig:data:assumption}) that both local and global features contribute to the classification but global features dominates the classification performance. 

At the same time, reducing the local information while preserving the global information enhances the property of \emph{minimality} of the representations, which is tested by the linear probe. Table \ref{tab:mnist} shows that, different from the decreasing nonlinear probe, the linear probe result increases when we increase the dependency horizon. \emph{This phenomenon gives a counter example of a hypothesis that raised in the previous literature~\cite{hewitt2019designing,qian2016investigating,belinkov2017neural}, which states the linear and nonlinear probes have similar trends.}

We also report the evaluations of a FPVAE in Table \ref{tab:mnist}. We can find that the results  are very close to the LPVAE with $h=2$. This  suggests that although the full autoregressive decoder 
has the ability to capture both local and global features, it still prefers to learn the local features during training and leaves the global features to be captured by the latent in practice.

\subsection{Representation Learning on CIFAR10} \label{cifar_exp} 
We conduct the same comparisons for CIFAR10, where we use a VAE with $\mathrm{Dim}(Z)=64$ and ResNet~\cite{he2016deep} with 3 convolutional  blocks in both encoder and decoder. The decoder's output has channel size 100 and is fed into a PixelCNN with 5 residual blocks  \cite{van2016pixel}.
For color pixels, observational distribution is a mixture of 10  logistic distributions with linear autoregressive  within channels \citep{salimans2017pixelcnn++}. Therefore, for a VAE with a conditional independent decoder, the independence is between super-pixels (each super-pixel contains 3 RGB channels). We also apply the pre-possessing method that used in \cite{ma2020decoupling}: random horizontal flipping and random cropping after padding with 4 pixels. All the models are trained using Adam~\cite{kingma2014adam} with $\text{lr}=3\times 10^{-4}$ for 1000 epochs.


\begin{table}[t]
 \caption{Representation learning on CIFAR10, both probe results are calculated over 3 random seeds. \label{tab:cifar}}
      \centering
 \begin{tabular}{cccccc}\hline
       Model & VAE ($h=0$) & LPVAE ($h=1$) & LPVAE ($h=2$) & FPVAE \\
        BPD & 4.98  & 3.57& 3.20& \textbf{3.03} \\ \hline
        Mutual Info & 185.3   & 108.66 &  98.47 & 47.51\\
        Intrinsic Dim & 64  & 62 & 62 & 48\\ \hline
        Nonlinear Probe& $57.90\pm 0.3$ &  $72.62\pm0.2$ & $73.16\pm0.1$ & \textbf{75.31}$\pm$0.1 \\
        Linear Probe & $46.82\pm 0.1$ &  $66.79\pm 0.2$ & $68.55\pm0.1$  & \textbf{71.10}$\pm$0.4 \\ \hline
\end{tabular}
\end{table}
\begin{figure}[t]
\vspace{-0.2cm}
\centering
    \begin{subfigure}[t]{0.24\linewidth}
    \centering
        \includegraphics[width=\textwidth]{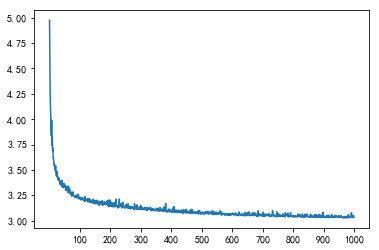}
        \caption{{\footnotesize BPD}}
    \end{subfigure}
    \begin{subfigure}[t]{0.24\linewidth}
    \centering
        \includegraphics[width=\textwidth]{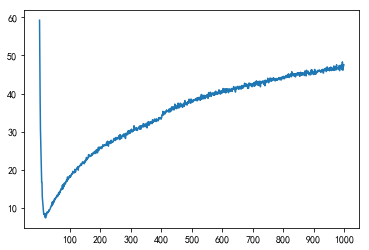}
        \caption{{\footnotesize Mutual Information}}
    \end{subfigure}
    \begin{subfigure}[t]{0.24\linewidth}
    \centering
        \includegraphics[width=\textwidth]{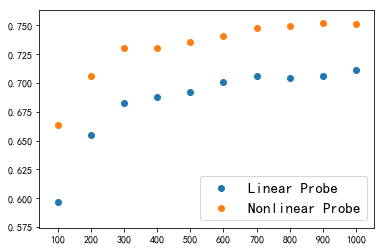}
        \caption{{\footnotesize Lin./Nonlin. Probes}}
    \end{subfigure}
    \begin{subfigure}[t]{0.24\linewidth}
    \centering
        \includegraphics[width=\textwidth]{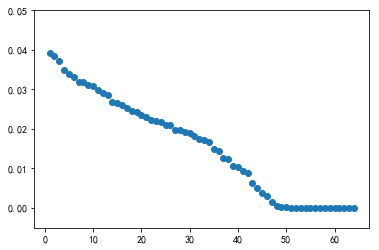}
        \caption{{\footnotesize Intrinsic Dimension}}
    \end{subfigure}
    \caption{In Figure a and b, we show the test BPD and mutual information after each training epoch. We plot the linear/nonlinear probe results for every 100 epochs in Figure c. Figure d shows the eigenvalues of the learned representations evaluated in the 1000th epoch. \label{fig:fpvae:dynamics}}
\end{figure}
Table \ref{tab:cifar} compares the representations learned by VAE variants. For CIFAR10, a conditional independent VAE is no longer flexible enough to model the data distribution well and only achieves 4.98 BPD. Therefore, a lot of  information is lost during training including both local and global features.  When we increase the dependency horizon of the PixelCNN, the decoder becomes  more  powerful to capture local features, which leads to the improvements of the BPD. Additionally, since more local features are captured in the decoder, less remaining information  is required to be captured by the latent, so we can find the FPVAE has the smallest mutual information and lowest intrinsic dimension (see also Figure \ref{fig:fpvae:dynamics}d) of the representations, which is consistent to the desired \emph{minimal} property. 

\begin{figure*}
\vspace{-0.2cm}
\centering
 \begin{subfigure}[h]{0.3\textwidth}
         \centering
         \includegraphics[width=\textwidth]{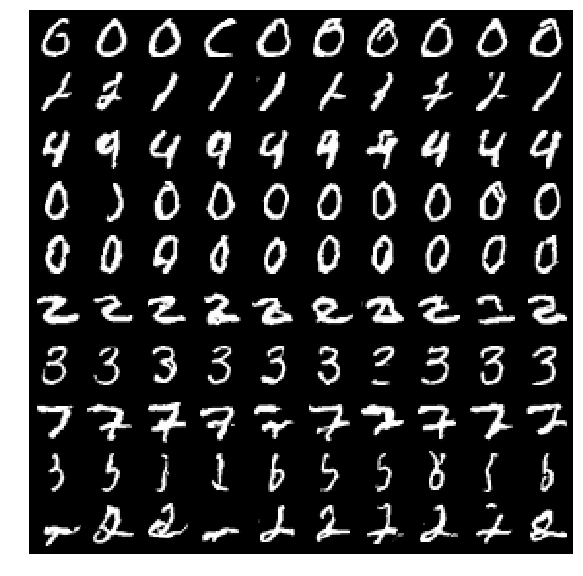}
     \end{subfigure}
 \begin{subfigure}[h]{0.5\textwidth}
         \centering
         \includegraphics[width=\textwidth]{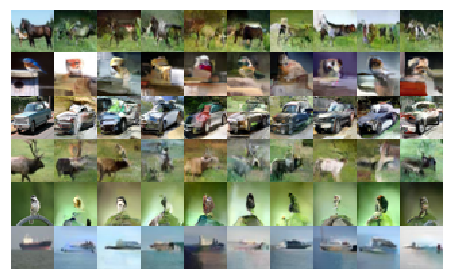}
     \end{subfigure}
     \caption{Samples from the FPVAE models that are trained on MNIST and CIFAR10. Images in the first column of both figures are true data samples, which we denote as $x'$. The remaining images in each row are generated using the same latent code $z'$ where  $z'\sim q_\phi(z|x')$. Therefore, all the images in each row shares the same latent $z'$. We can find samples in a row shares the same global features, but the local features are different from each other. \label{fig:local:gen}}
     \vspace{-0.3cm}
\end{figure*}

To understand the learning dynamics of FPVAE, we  plot the trends of the test BPD, mutual information and linear/nonlinear probes during training, see Figure \ref{fig:fpvae:dynamics}.
 We can find in the beginning of training, the BPD quickly drops to 3.2 and the mutual information also drops below 5, which is very close to the latent collapse phenomenon. In this case, the decrease of the BPD is mainly contributed by the PixelCNN decoder and the latent doesn't learn too much information about the data.
\begin{wrapfigure}{r}{0.6\linewidth}
\vspace{-0.0cm}
  \centering
  \begin{minipage}[c]{0.49\linewidth}
\begin{center}
 \begin{tabular}{c c} 
 \hline
Model & Accuracy  \\ [0.5ex] 
 \hline
Raw Image$^\ddag$& 35.32\\
 \hline
AAE$^\dagger$& 37.76 \\ 
VAE$^\dagger$ &39.59\\
NAT$^\dagger$ & 39.59 \\
BiGAN$^\dagger$ & 44.90 \\
DIM(G)$^\dagger$ & 29.08\\
DIM(L)$^\dagger$ & 49.62\\
FlowVAE$^\ddag$ & 59.53  \\
 \hline
FPVAE & \textbf{71.10} \\ 

 \hline
\end{tabular}
\subcaption{Linear probe}
\end{center}
\end{minipage}
 \begin{minipage}[c]{0.49\linewidth}
\begin{center}
 \begin{tabular}{c c} 
 \hline
Model & Accuracy  \\ [0.5ex] 
 \hline
Supervised$^\dagger$ & 75.39\\ 
 \hline
VAE$^\dagger$ &54.61\\
$\beta$-VAE$^\dagger$& 55.43\\
AAE$^\dagger$ & 52.81 \\ 
BiGAN$^\dagger$ & 52.54 \\
DIM (DV)$^\dagger$ & 64.71\\
DIM (JSD)$^\dagger$ & 66.96\\
DIM (NCE)$^\dagger$& 69.13\\
\hline
FPVAE& \textbf{75.31}  \\
 \hline
\end{tabular}
\subcaption{Nonlinear probe}
\end{center}
\end{minipage}
\captionof{table}{Linear and nonlinear classification accuracy comparisons. Results with $\dagger$ are  from \cite{hjelm2018learning} and  $\ddag$ are from \cite{ma2020decoupling}. \label{tab:cifar:pretraining}}
\vspace{-0.2cm}
\end{wrapfigure}
 This phenomenon is consistent to the assumption that local features dominates the BPD \cite{schirrmeister2020understanding,zhang2021out}. However, when we train the model for a longer time (from 100 to 1000 epochs),  the mutual information starts to increase as well as the linear/nonlinear probes, which indicates the latents start to learn global information that is related to the labels. During this process, the BPD only has a marginal decrease $\approx 0.2$, which suggests although global features are the key to the downstream classification, they contribute much less to the BPD comparing with  the local features. This phenomenon also explains why the likelihood based models are usually not very competitive in the representation learning \cite{hjelm2018learning}. In Figure \ref{fig:local:gen}, we also  show the samples from the FPVAE to help visualize the decomposition of the local and global features.
In Table~\ref{tab:cifar:pretraining}, we compare the FPVAE with other methods including VAE~\citep{kingma2013auto}, $\beta$-VAE~\citep{higgins2016beta}, AAE~\citep{makhzani2015adversarial}, BiGAN~\citep{donahue2016adversarial}, NAT~\citep{bojanowski2017unsupervised}, Deep InfoMAX (DIM)~\citep{hjelm2018learning} and FlowVAE~\citep{ma2020decoupling}. We can find FPVAE significantly outperform other methods in both linear and nonlinear probes. 

\subsection{Relation to FlowVAE}
The most related work on improving VAE-based representation learning is the recent proposed FlowVAE~\citep{ma2020decoupling}, which uses a flow  as a part of the  decoder $ p_\theta(x)=\int \delta(x-f_\theta(g_\theta(z),v)) p(z)p(v)dzdv$,
where the $g$ is a neural network maps from low-dimensional $z$ to a space that has dimension equal to $x$ 
and $f$ is a invertible flow function and the $v$ is also a latent variable whose dimension is the same as the $x$. 
The model is shown empirically to be able to decouple local and global features and can improve the representation learning results by using the ``global'' latent variable $z$. 
This phenomenon can also be explained by a  hypothesis proposed by \cite{kirichenko2020normalizing}: \emph{Flow layers learn generic
image-to-latent-space transformations that leverage local pixel correlations and graphical
details rather than the semantic content}. 
This suggests the shallow flow decoder in the FlowVAE prefers to learn the local features so that the latent representation $z$ can capture global features.
Although this motivation is similar to ours,
we argue that using a PixelVAE-style model 
allows a more transparent 
\begin{wrapfigure}{r}{0.5\textwidth}
\centering
\vspace{-0.2cm}
\captionof{table}{BPD comparison with FlowVAE~\cite{ma2020decoupling}.\label{tab:BPD:flowvae}}
\begin{tabular}{c c c c} 
 \hline
Model &  FlowVAE & LPVAE (h=2) &  FPVAE  \\
 \hline
 BPD & 3.27 & 3.20 & \textbf{3.03}\\
  \hline
\end{tabular}
\end{wrapfigure}
study of the learning behavior
(e.g. the dependency horizon can be controlled) and leads to better representation learning performances as shown in Table \ref{tab:cifar:pretraining}. 
Additionally, PixelVAE-style models can also achieve higher BPD comparing to FlowVAE, see Table \ref{tab:BPD:flowvae}.

\section{Semi-Supervised Learning}
Another application of  representation learning is the semi-supervised learning \cite{bengio2013representation,kingma2014semi}, where the training set contains both labeled data $\mathcal{X}^l=\{(x^l_1,y_1),\ldots,(x^l_N,y_N)\}$  and unlabeled data $\mathcal{X}^u=\{x^u_1,\ldots,x^u_M\}$. For $\mathcal{X}^l$, we can build a joint model $p_\theta(x,y)$ to the data. For $\mathcal{X}^u$, a uniform prior can be placed over the classes and
$p_\theta(x^u)=\frac{1}{K}\sum_k p_\theta(x^u,y)$ can be used to fit the data.  For VAE models, both log-likelihood function are replaced by their lower bounds for training. The lower bound of $\log p_\theta(x^l,y)$ is an simple extension of the standard ELBO 
\begin{align}
 \log p_\theta(x^l,y)&\geq \langle\log p_\theta(x^l,y,z)-\log q_\phi(z|x^l,y)\rangle_{q_\phi(z|x^l,y)}\equiv \mathrm{ELBO}(x^l,y)\label{eq:elbo:joint}.
\end{align}
For the unlabeled data model $p_\theta(x^u)$, \cite{kingma2014semi} proposed the following lower bound:
\begin{align}
    \log p_\theta(x^u)&\geq \left\langle\log p_\theta(x^u,y,z)-\log q(z,y|x^u)\right\rangle_{q_\phi(z,y|x^u)}\equiv \mathrm{ELBO}(x^u), \label{eq:elbo:unlablled}
\end{align}
 where they introduce an additional classifier (ac) with parameter $\psi$: $q_\phi^{ac}(y|x^u)$ to construct the variational distribution
$q_{\phi,\psi}(z,y|x^u)=q_\phi(z|x^u,y)q_\psi^{ac}(y|x^u)$.
In practice, it is useful to add a cross-entropy regularizer into the training objective \ref{eq:semi}, the final objective is then
\begin{align}
&\frac{1}{N}\sum_{n=1}^N \mathrm{ELBO}(x_n^l,y_n)+\frac{1}{M}\sum_{m=1}^M \mathrm{ELBO}(x_m^u)+\alpha  \frac{1}{N}\sum_{n=1}^N \log q_\psi^{ac}(y_n|x^l_n).\label{eq:semi}
\end{align}
This framework is referred to as \emph{M2} model~\cite{kingma2014semi}. In practice, $\alpha$ is chosen to be $r\frac{N+M}{M}$, where $N$ and $M$ are the sizes of the labeled/unlabeled datasets and $r$ is the supervision rate. We follow \cite{kingma2014semi,siddharth2017learning} to choose $r=0.1$ in all our experiments. We use LPVAE with $h=1$ and $h=2$ for MNIST and SVHN respectively, the latent dimension is 64 in both cases. For MNIST experiments, we use a VAE with both encoder and decoder contains  a three layers fully connected networks with ReLU activations. The output of the decoder is further fed into the PixelCNN module as described in  Section \ref{mnist_exp}. The classifier is a three layer fully connected network with ReLU activations. The model is trained with 50 epoch using batch size 16.  For SVHN experiments, we use a VAE with  the encoder has the architecture of four convolutional layers, each with kernel size 5 stride 2 and padding 2, and two fully connected layers as well as using batch normalization and leaky ReLU for activations.   Likewise, the decoder has two fully connected layers, 4 transposed convolutional layers as well as using batch normalization and leaky ReLU for activations. The autoregressive module has the same setting as Section \ref{cifar_exp}. The classifier contains 2 convolutional layers and 3 fully connected layers with batch normalization and ReLU as activations. Dropout is also used in the classifier with 0.3 dropout rate. The model is trained for 20 epoch with batch size 32.

Different from the unsupervised pre-training task, the representations are now learned jointly with the class labels. Our goal is to show that by using a decoder that can learn local features, the remain global features can be well-captured by the representation and thus improves the data efficiency. For MNIST experiments, we split the training data into labeled and unlabeled dataset and varies the labeled data from 100 (10 per class) to 3000 (300 per class). Similarly, we vary the labeled data number from 100 to 3000 for SVHN~\cite{netzer2011reading}. 
\begin{table}[h]
\small
\centering
\caption{Semi-supervised learning comparisons for  MNIST and SVHN. \label{tab:semi}}
\begin{center}
 \begin{tabular}{ccccccccc} 
  \toprule
  &\multicolumn{4}{c}{MNIST} & \multicolumn{2}{c}{SVHN}\\
  \cmidrule(lr){2-5} \cmidrule(lr){6-7}
Models & 100 & 600 & 1000 & 3000 &1000&3000  \\ 
\hline
M2   & 88.03($\pm 1.7$) &  95.06($\pm 0.1$) & 96.40($\pm 0.6$)& 96.08($\pm 0.6$) &63.98$\pm$0.1 & -\\ 
EQVAE  & 91.10($\pm 0.7$) & 96.01($\pm 0.2$)  & 96.66($\pm 0.2$)  & 97.77($\pm 0.1$) & 62.05$\pm$0.7  & 75.05$\pm$0.6\\
DisVAE  & 90.29($\pm 0.9$) & 96.16($\pm 0.9$) & 97.12($\pm 0.8$) & \textbf{98.43}($\pm 0.9$) & 61.09$\pm$1.0 & 70.93$\pm$0.8\\
Ours  & \textbf{96.41}($\pm 0.4$) & \textbf{97.08}($\pm 0.2$)  & \textbf{97.15}($\pm 0.2$) & 97.46($\pm 0.1$)&\textbf{80.17}$\pm$ 1.2 & \textbf{84.48}$\pm$0.3 \\
\hline
\end{tabular}
\end{center}
\end{table}


\begin{wrapfigure}{r}{0.5\textwidth}
\vspace{-0.5cm}
\captionof{table}{Comparison with FlowGMM \citep{izmailov2020semi}.}
\begin{tabular}{c c c} 
 \hline
Models & MNIST (1k) &  SVHN (1k)\\
 \hline
Ours & 97.15$\pm$ 0.2 & 80.17$\pm$1.2 \\
 FlowGMM  & 98.94 & 82.42\\
  \hline
\end{tabular}
  \vspace{-0.1cm}
\end{wrapfigure}

In Table \ref{tab:semi}, we report the comparisons with other VAE-based  semi-supervised methods:  VAE M2/M1+M2~\citep{kingma2014semi}, EQVAE \citep{feige2019invariant}, DisVAE \citep{siddharth2017learning}.
We can find LPVAE outperforms other VAE variants in most cases, especially when the label number is limited. This shows  our model can also improve the data efficiency when the representations are jointly learned with the task. 
For reference, we also report the comparison with the SOTA likelihood-based semi-supervised models: FlowGMM \citep{izmailov2020semi}. We can see FlowGMM is slighter better than LPVAE in the conducted experiments. However, since flow  models don't allow a low-dimensional representation, these two models are not directly comparable for the purpose of representation learning.

\section{Conclusion}
In this work, we conducted a comprehensive study of the VAE-based representation learning. We have shown that, by incorporating the right inductive bias into the model design, we can significantly improve the representation quality and benefit both down-stream tasks and semi-supervised learning.


\bibliography{ref}
\bibliographystyle{abbrv}

\end{document}